\documentclass[10pt,twocolumn,letterpaper]{article}

\usepackage{iccv}
\usepackage{times}
\usepackage{epsfig}
\usepackage{graphicx}
\usepackage{amsmath}
\usepackage{amssymb}
\usepackage{subfig}
\usepackage{textcomp}
\usepackage{multirow}
\usepackage[accsupp]{axessibility}
\usepackage[pagebackref=true,breaklinks=true,letterpaper=true,colorlinks,bookmarks=false]{hyperref}

\iccvfinalcopy % *** Uncomment this line for the final submission

\def\iccvPaperID{2599} % *** Enter the ICCV Paper ID here

\ificcvfinal\fi

\begin{document}

%%%%%%%%% TITLE
\title{What You Can Learn by Staring at a Blank Wall}

\author{Prafull Sharma$^{1}$
\qquad
Miika Aittala$^{1,2}$
\qquad
Yoav Y. Schechner$^{3}$
\qquad
Antonio Torralba$^{1}$
\and
Gregory W. Wornell$^{1}$
\qquad
William T. Freeman$^{1}$
\qquad
Fr\'edo Durand$^{1}$\\
\\
$^{1}$MIT \quad $^{2}$NVIDIA \quad $^{3}$Technion - Israel Institute of Technology
}

\maketitle
\ificcvfinal\thispagestyle{empty}\fi

%%%%%%%%% ABSTRACT
\begin{abstract}
    We present a passive non-line-of-sight method that infers the number of people or activity of a person from the observation of a blank wall in an unknown room. Our technique analyzes complex imperceptible changes in indirect illumination in a video of the wall to reveal a signal that is correlated with motion in the hidden part of a scene. We use this signal to classify between zero, one, or two moving people, or the activity of a person in the hidden scene. We train two convolutional neural networks using data collected from 20 different scenes, and achieve an  accuracy of $\approx94\%$ for both tasks in unseen test environments and real-time online settings.
    Unlike other passive non-line-of-sight methods, the technique does not rely on known occluders or controllable light sources, and generalizes to unknown rooms with no re-calibration. We analyze the generalization and robustness of our method with both real and synthetic data, and study the effect of the scene parameters on the signal quality.\footnote{Code, data, and video available at \href{http://wallcamera.csail.mit.edu}{wallcamera.csail.mit.edu}.}
\end{abstract}

%%%%%%%%% BODY TEXT
\section{Introduction}

Consider a situation where one would like to recover information about a hidden scene in an unknown room without directly peeking inside. Staring at the blank wall of the room from outside may reveal nothing to the naked eye, yet the wall reflects extremely faint but meaningful patterns of light from the hidden scene. We show that by analyzing a video of the blank wall, we can infer information about a person's activity or classify the number of people in a hidden region of the scene, with no prior calibration or knowledge of the environment. Real-time in-situ use of such a non-line-of-sight (NLOS) method can be critical for search and rescue operations, law enforcement, emergency response, fall detection for the elderly, and detection of hidden pedestrians for intelligent vehicles \cite{borges2012pedestrian,naser2018shadowcam}.

\begin{figure}
    \centering
    \includegraphics[width=\columnwidth]{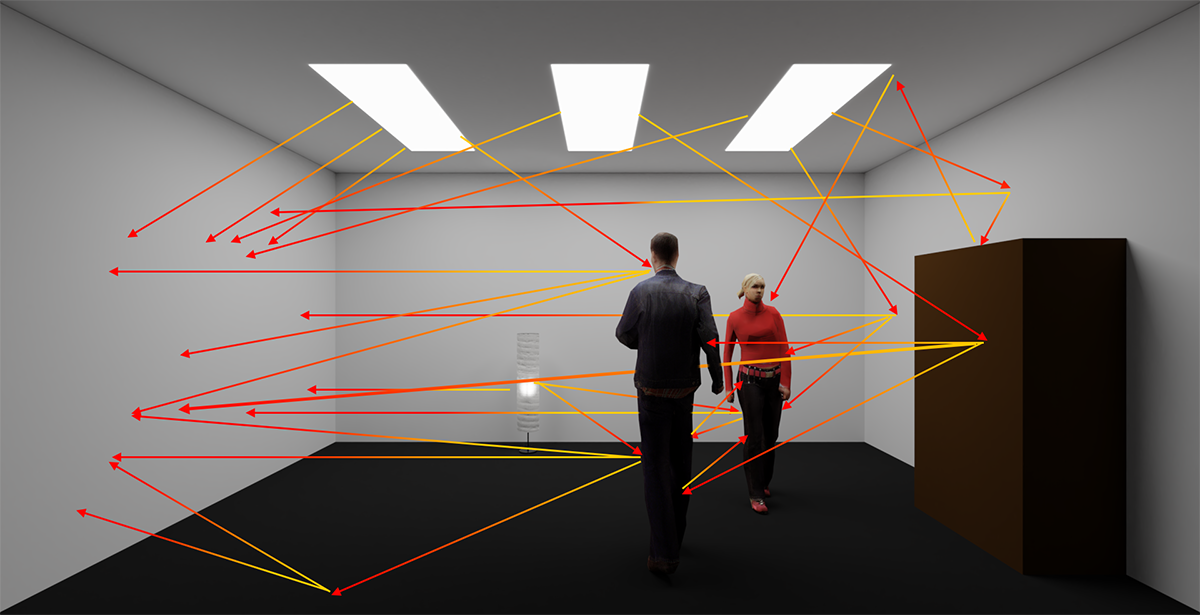}
    \caption{The light originating from the lamps takes infinitely many paths to reach the wall on the left. Motion of people in the room changes the mutual visibility between the wall and the rest of the scene, which induces subtle changes in patterns of light observed on the wall. Notice that there are no direct shadows, as the persons do not occlude the direct paths connecting the left wall and the lamps.}
    \label{fig:lightpaths}
\end{figure} 

% How is this novel? How are you different from other techniques
NLOS imaging techniques have previously explored related imaging setups using both {\em active} and {\em passive} methods. Active methods interact with the environment using flashlights, projectors, lasers, WiFi signals, or sound to extract information about the hidden scenes \cite{Adib:2015:CHF:2816795.2818072,adib2013see,chandran2019adaptive,depatla2015occupancy,heide2017non,kadambi2016occluded,klein2016tracking,Lindell:2019:Acoustic,metzler2019keyhole,velten2012recovering}. Removing the requirement for probing devices, passive approaches only use cameras, without controlling the lighting conditions or interacting with the environment in any manner. Existing passive methods typically exploit known occluders such as a visible corner or a hidden table fan, which act as accidental imaging devices \cite{aittala2019computational,baradad2018inferring,bouman2017turning,saunders2019computational,seidel2020two,seidel2019corner,Tanaka_2020_CVPR,tancik2018flash,tancik2018data,yedidia2019using}. In such work, the occluders provide structure to the light transport, enabling reconstruction of human-readable images and tracks of motion in the hidden scene from the observed indirect illumination.

In contrast, we do not rely on occluders between the scene of interest (the persons) and the observed blank wall. Instead, we leverage the complex temporal variation of indirect illumination between a person, the room, and potentially a second person. We use a video of the blank wall and show that a learning-based approach is able to recover information about the hidden scene. Specifically, we demonstrate automatic classification of the number (between zero, one, and two) and activity of the moving people. 
Figure \ref{fig:lightpaths} illustrates light transport~\cite{Kajiya:1986:RE:15922.15902} occurring in a typical uncontrolled room environment with multiple light sources, objects, and people. The seemingly featureless illumination on the left wall is in fact a complex mixture of light reflected along different paths in the scene.
The dominant component is the direct illumination from the light sources themselves, but a small fraction is contributed by light bouncing from the objects in the hidden scene, and yet smaller fraction from light that has bounced multiple times.
The pattern of light cast onto the wall depends on a complex interplay of the mutual visibility of the geometry and the materials in the hidden scene.
People moving in the scene affect light paths, creating new interreflections and blocking other contributions, which leads to subtle temporal variations in the light reaching the wall.
In a typical video imaging setup, the magnitude of the temporally changing illumination is extremely low compared to the imaging noise: in our measurements, often $-20$ dB to as low as $-35$ dB. 
The spatial characteristics of these variations are also elusive as they depend on the scene, the illumination, and the relative position of the persons.

Despite the extremely low signal levels and high noise in the observed videos of the wall, we extract a signal that summarizes the essential motion in the hidden scene by projecting the video into a 2D representation 
(Section \ref{section:signal_extraction}). We use this 2D representation of the video to train convolutional neural networks to perform two tasks: classifying the activity of a person between walking, jumping, waving hands, crouching, and no activity (all objects being static), and inferring whether zero, one, or two people are present in the hidden scene (Section \ref{section:method}). Our models trained on 20 different scenes achieve an accuracy of $94.4\%$ in classifying the number of people and $93.7\%$ in activity recognition on the held out test set of 5 unseen scenes. Our method works in real time with almost no latency without any scene specific calibration. We further investigate the impact of the properties of the scene and human motion on this task using a theoretical framework and a synthetic model of the setup (Section \ref{section:results}).

\section{Related Work} \label{section:related_work}

\textbf{Active methods.} Many active methods use time-resolved detectors in an NLOS setup for shape prediction \cite{adib20143d,adib2013see,heide2014diffuse,heide2017robust,kirmani2009looking,lindell2019wave,o2018confocal,xin2019theory,Rapp2020SeeingAC,metzler2019keyhole}, tracking \cite{adib20143d,chan2017non,gariepy2016detection,Scheiner_2020_CVPR,wu2021non}, and person identification \cite{caramazza2018neural}. Adib et al. have used WiFi signals to infer human shape behind walls and to predict 3D trajectories of the hands of the people in the hidden scene \cite{Adib:2015:CHF:2816795.2818072,adib2013see}. WiFi signals have also been used to predict the number of people in the hidden scene \cite{depatla2015occupancy}. NLOS data collected with pulsed laser arrays and single photon array detectors have been used for person identification with neural networks \cite{caramazza2018neural}. Metzler et al. use pulsed laser and time-resolved detector to predict the shape and trajectory of the hidden object \cite{metzler2019keyhole}. Picosecond laser and SPAD detectors have recently been used to recover the albedo and shape of objects in the hidden scene \cite{o2018confocal}.

Some active methods also perform the similar NLOS tasks by interacting with the environment using steady-state intensity sources (without time resolved data) and a regular camera \cite{chandran2019adaptive,chen2019steady,gupta2012reconstruction,klein2016tracking,metzler2020deep,smith2018tracking}. Klein et al. propose a method to track objects in the hidden scene using a laser pointer and a camera pointed at a blank wall \cite{klein2016tracking}. Other methods use a projector and conventional camera to perform 3D localization and identification of objects in the hidden scene \cite{chandran2019adaptive}.

Active methods can recover useful information about activity in the hidden scene, but may not be feasible (due to power consumption, time, and complexity of the measurements) or desirable for certain applications. This makes it hard to deploy them outside of the laboratory setting.

\textbf{Passive methods.} Many passive techniques rely on external occluders to form an accidental image of the scene of interest. Recent passive methods can recover the position of an occluder with a known shape and a planar image in the hidden scene behind it from a single photograph \cite{saunders2019computational}. Other works recover a lightfield from shadows cast by an occluder of known shape between the scene and the observation plane \cite{baradad2018inferring} or recover video of the hidden scene given a complex occluded observation scene \cite{aittala2019computational}. Tancik et al. present a per-scene VAE model for reconstructing the hidden part of the room by observation of a flat floor and posts on the floor \cite{tancik2018data}. Another work demonstrates a data driven method for localization, identification, and hidden scene reconstruction for a known scene geometry and controlled illumination conditions (flashlight) \cite{tancik2018flash}. The above methods are restricted to the laboratory as they assume knowledge about the scene geometry, illumination conditions, or objects in the scene. We present a method that recovers useful information in less constrained environments.

Bouman et al. \cite{bouman2017turning} use corner occluders to recover a 1D angular projection of the hidden scene by exploiting faint penumbra cast on the ground near a corner. Visualizations of the 1D angular projections allow a human interpreter to predict the number of people or to guess their activity.
Occluders are also exploited by Torralba and Freeman to form accidental cameras \cite{TorralbaAccidental}, revealing 2D images from pinhole or pinspeck occluders. Both methods operate outside of the laboratory, but require a static occluder in the environment, and return a 1D or 2D image to be interpreted by an observer. Our method removes the need for an occluder in the environment, and just needs a blank wall, visible to both the camera and the hidden scene. Estimates of numbers of people, and their activities, are returned automatically.
\section{Imaging Setup}\label{section:overview}

\begin{figure}
    \centering
     \subfloat[\label{fig:shadow}]{
      \includegraphics[width=0.5\columnwidth]{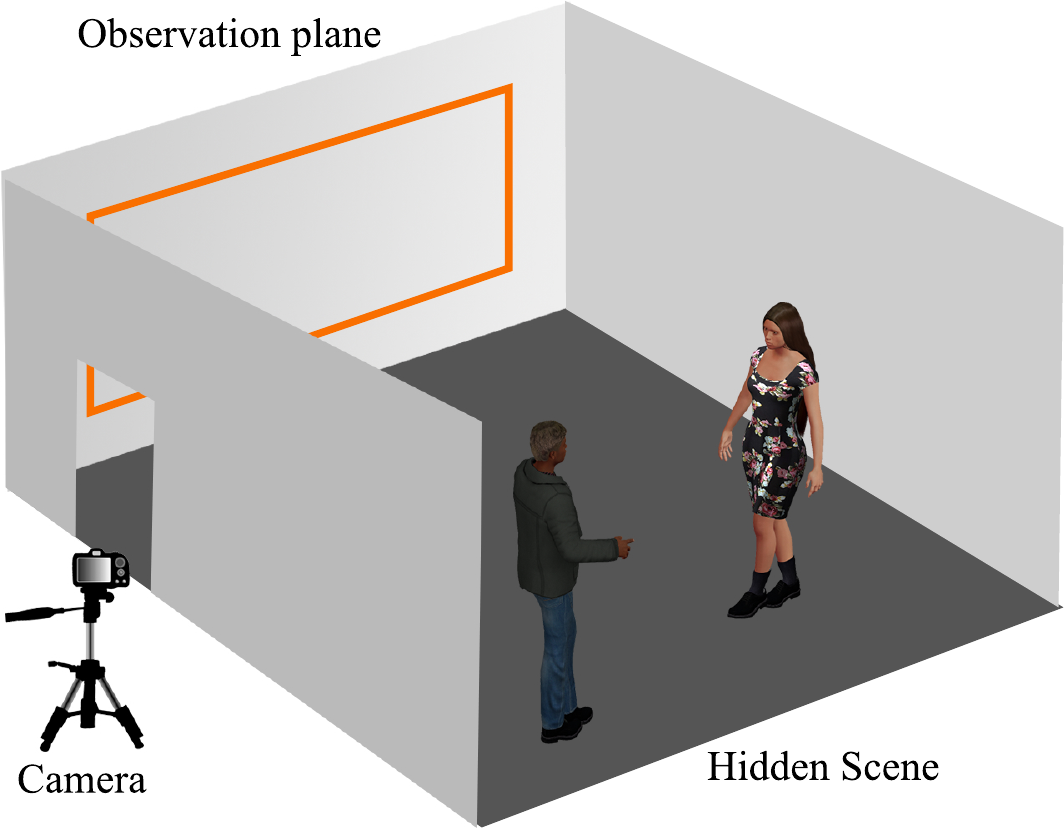}
     }
     \subfloat[\label{fig:reflection}]{
      \includegraphics[width=0.45\columnwidth]{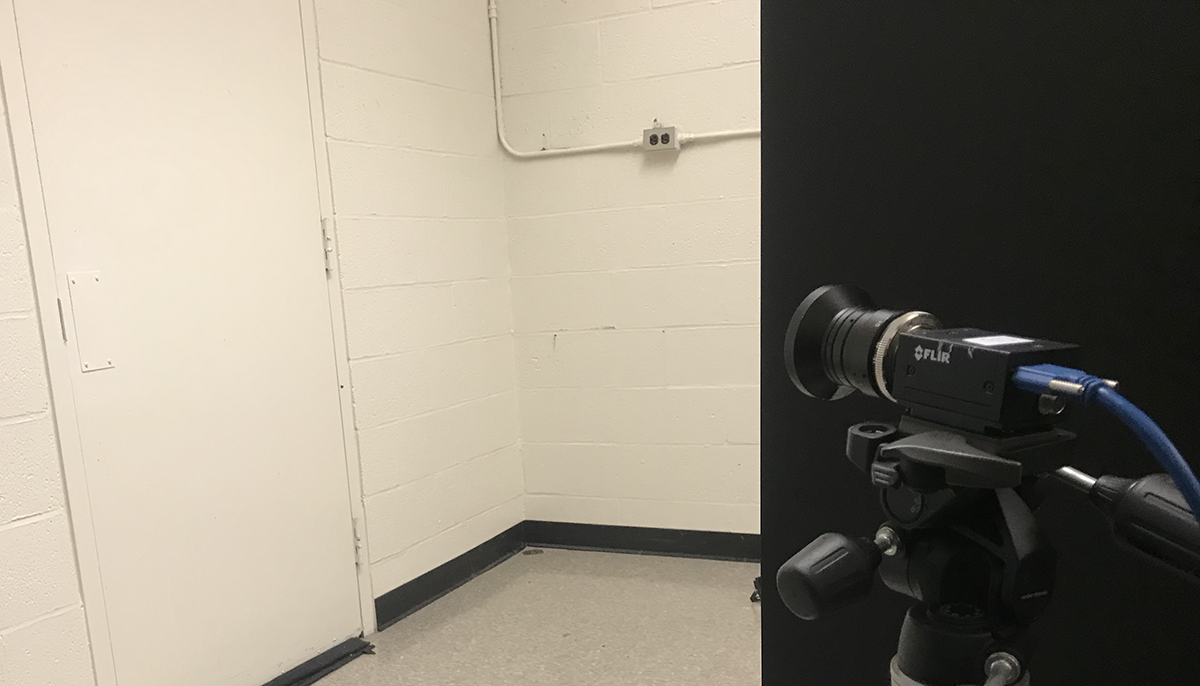}
     }
     \caption{(a) Setup of a possible scenario with two people walking in the hidden scene with the camera observing the wall from outside the room. (b) Camera observing the wall in a real scenario in one of our scenes.}
     \label{fig:setup}
\end{figure}

We use a camera outside a room to record video of a blank wall inside the room, while people move in the hidden scene (Figure \ref{fig:setup}). Each point on the observed wall receives light directly from the light sources and from the different objects in the room after multiple bounces, as presented in Figure \ref{fig:lightpaths}. Under typical indoor illumination conditions, the observed wall would appear static to the naked eye, even though there might be activity in the hidden scene. This is due to the dominant contribution of the \emph{ambient light} coming directly from the light sources and reflections from static objects. However, the light bouncing off moving objects in the room have a subtle effect on the observed wall.

Movement of people in the hidden scene blocks and reflects parts of the indirect light within the hidden scene, causing small temporal changes on the pattern of light observed on the wall. The changes depend on a variety of unknown factors, such as relative positions, colors and intensities of the hidden objects and light sources. Similar activities can induce very different patterns in different rooms. As the person is small compared to the room dimensions, their contribution is extremely small as compared to the ambient light hitting the wall -- often below $1\%$ of the maximum pixel intensity and below the noise level of the video, corresponding to a signal-to-noise ratio as low as $-35$ dB.

\section{Signal Extraction and Processing} \label{section:signal_extraction}
Our method begins by extracting the subtle motion signal from the raw input video, and then further projects it into a 2D space-time plot representation which is then used as an input for classification.

%%%%%%%%%%%%%%%%%%%% FIGURES %%%%%%%%%%%%%%%%%%%%%%%%%%%%%%%
\begin{figure}[t]
\centering
\includegraphics[width=0.9\columnwidth]{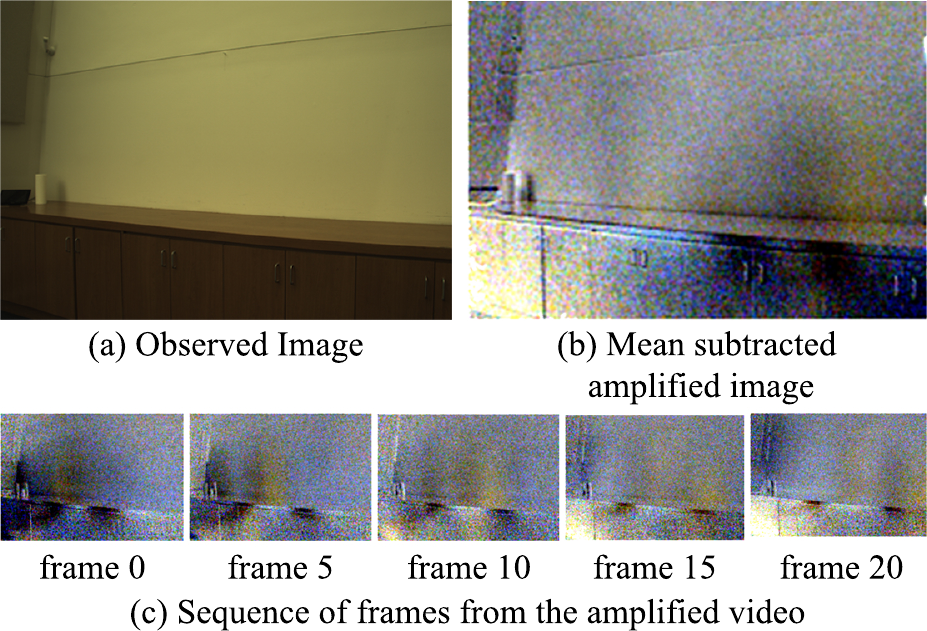}
\caption{(a) A representative frame of the seemingly static input video. (b) A frame of the amplified residual video after subtracting the mean frame reveals faint changes in illumination caused by motion of the people. (c) A sequence of frames shows the motion of these features.}
\vspace{-1em}
\label{fig:meansubsignal}
\end{figure}
\begin{figure*}
\centering
\includegraphics[width=0.7\textwidth]{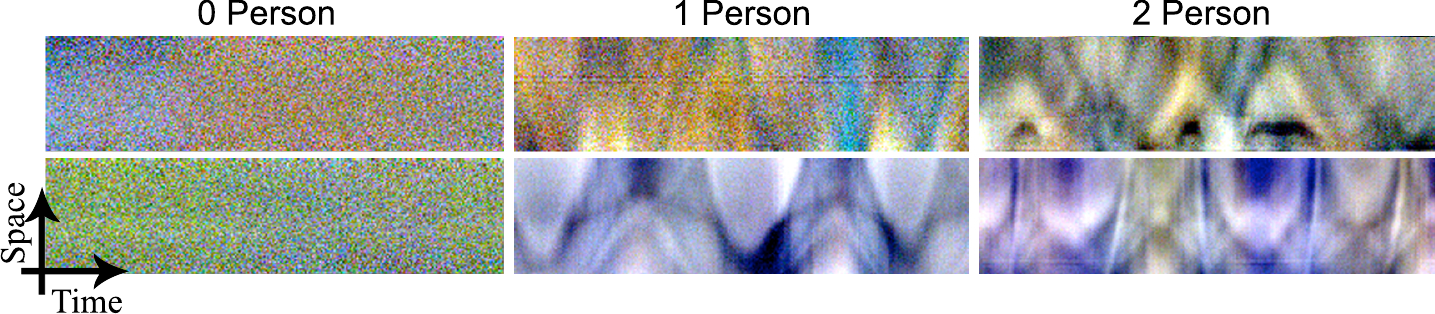}
\caption{Examples of horizontal space-time plots for zero, one, and two person(s) cases. The space-time plots exhibit different visual features for each class, enabling classification using a convolutional neural network.}
\vspace{-0.5em}
\label{fig:stplot}
\end{figure*}
\begin{figure*}
\centering
\includegraphics[width=\textwidth]{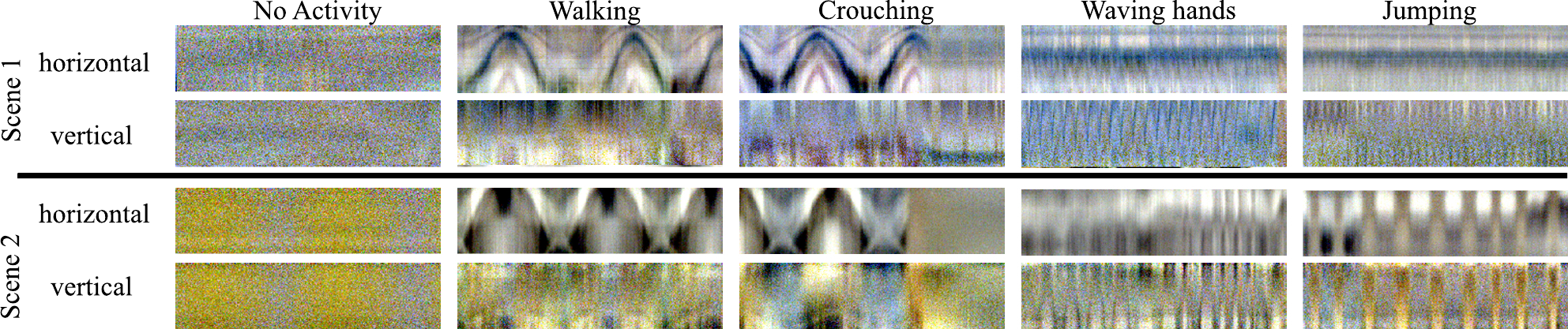}
\caption{Examples of horizontal and vertical space-time plots for activity recognition. See video of real-time demonstration at \href{http://wallcamera.csail.mit.edu}{wallcamera.csail.mit.edu}.
}
\vspace{-1em}
\label{fig:actrecog_stplot}
\end{figure*}
%%%%%%%%%%%%%%%%%%%%%%%%%%%%%%%%%%%%%%%%%%%%%%%%%%%%%%%%%%%%%

\subsection{Signal Extraction}
As discussed in Section \ref{section:overview}, the recorded video of the wall is dominated by temporally unchanging ambient light whose magnitude obscures the subtle dynamic signal from the moving persons.
We approximately remove this constant light by computing the temporal average of the video, and subtracting it from each frame, as suggested by previous work on accidental pinholes \cite{TorralbaAccidental}. Note that the pixels can now also have negative values. The resulting video reveals the temporal changes in illumination pattern caused by the motion of the persons.

The magnitude of the residual signal is typically very small compared to the original pixel values, and often far below the noise level of the video. To reduce the noise, we spatially downsample the video by a factor of 16 ($4 \times$ in each dimension). We linearly scale the resulting video to the full range of RGB values by multiplying it by a factor of 50, and adding a middle-gray base level to make the negative values visible. This procedure results in the {\em amplified} video, illustrated in Figure \ref{fig:meansubsignal}. We encourage the reader to view the video in the supplementary material.

We also found that the signal of interest in amplified videos was sometimes overpowered by long-range periodic intensity sway, which we attribute to small discrepancies between the frame rate and the flickering of lights in the room at the AC frequency (60 Hz). To eliminate this effect, after the mean-subtraction we compute the median value of each frame (global 1D intensity curve over time) and project every pixel's temporal values to the orthogonal complement of this median curve.

Prior to any processing, we convert the input images to logarithmic space to equalize the magnitude of variations between light and dark regions of the scene. In case of extraneous objects in the observed video, we crop to the region of the blank wall.

\subsection{Space-Time Plots for Classification}
To further improve the signal-to-noise ratio and reduce the dimensionality to aid the learning process, we average the data along appropriate spatial dimensions of the video. Averaging the input data leads to noise reduction of approximately $45~ \mathrm{dB}$ while retaining the relevant motion signal; refer to Section \ref{section:snr} for analysis on signal-to-noise ratio.

For example, in the case when people are moving along the width of the observation plane, we average the video images over the vertical axis to obtain a 2D representation (space $\times$ time) that captures the essential spatial dimension of the motion.
This collapses the video ($W \times H \times T \times 3$) into a space-time plot of dimensions $W \times T \times 3$. Similarly, in activity recognition we also produce a space-time plot along the height of the observation plane, which captures the vertical motion such as crouching and jumping. We standardize the intensity and contrast of all space-time plots by clipping the $2\%$ of brightest and darkest pixels to discard outliers and normalizing to range $[0,1]$.

A related method for generating space-time plots is proposed by Bouman et al. \cite{bouman2017turning}.
Their plots show distinct visually interpretable tracks corresponding to the motion of the persons, whereas our space-time plots appear convoluted due to the absence of a known occluder. 

Figure \ref{fig:stplot} shows examples of horizontal space-time plots for zero, one, and two person(s) moving in the hidden scene, used for classifying the number of people. Note that the space-time plots for the three classes exhibit class specific visual features. The zero person space-time plots show almost no variation except for noise. One person plots show a blurry single track while space-time plots for the two person case show more complicated textures. Another notable characteristic in two person plots is a sharp vertical line that appears on the plot when the persons cross one another. This effect is apparent in Figure \ref{fig:stplot}. This sudden change in intensity of the wall appears when one person obscures the other, effectively making the scene resemble a one person scenario for a brief moment. 

In Figure \ref{fig:actrecog_stplot}, we show examples of the horizontal and vertical space-time plots used in the activity recognition task. We can observe distinct features corresponding to each dimension. For example, for walking the signal changes mostly in the horizontal axis, whereas for other activities there is signal in both directions, which is visible in the corresponding space-time plots. For crouching, the initial part of the space-time plot show features similar to that of walking followed by no motion, indicating a stationary crouching position. In jumping, there is signal in both the horizontal and vertical space-time plots, with rapid repetitive motion in both directions. The no activity category is the same as the zero person case as described for classifying the number of people. Motions with insufficient magnitude, e.g. small jumps in the same spot, can lead to similar space-time plots as the no activity scenario. Furthermore, it is easy to confuse between waving hands and jumping just by visually inspecting these space-time plots.

\section{Classification Method} \label{section:method}
Since the space-time plots for the various scenarios capture the essential motions in the spatial dimensions, this representation is suitable for learning to classify the scenarios using a convolutional neural network model. We use the space-time plots to solve two tasks, (i) classifying the number of people in the hidden scene and (ii) activity recognition. In this section, we describe the convolutional neural network architectures for these tasks.
\subsection{Classifying Number of People}
We want to classify between zero, one, or two person(s) moving in the hidden scene. We use the horizontal space-time plots (video averaged over the vertical dimension) as it captures the motion of the people along the horizontal dimension of the wall. The input to the network is an RGB space-time plot of dimension $64 \times 256 \times 3$, 64 pixels in spatial resolution and 256 time steps. The input to the network is standardized to zero mean and unit standard deviation.

Our neural network consists of 5 convolutional downsampling blocks, followed by max-pooling over time, and two fully connected layers which output the prediction. The convolutional layers extract spatio-temporal features and gradually bring the size of the feature maps down to the spatial resolution of $1 \times 13$ with $64$ feature channels. This represents a summary of spatio-temporal features at 13 local time instances. The pooling operation yields a single $64$-channel feature vector that aggregates local findings from different time instances into a joint statistic.
These are decoded by the final two layers into a predicted class.

\subsection{Activity Recognition}
We seek to classify between five activities: walking, jumping, walking followed by crouching, waving hands, and no activity. Since these activities have different temporal signatures along the horizontal and vertical directions, we use both horizontal and vertical space-time plots each of dimension $64 \times 256 \times 3$ for the classification task. 

The CNN architecture is identical to the network for classification of number of people, except that it contains two convolutional branches that process the two input space-time plots separately. The feature vectors from both branches are concatenated after the pooling layer, followed by the fully connected layers to predict the activity.

\subsection{Implementation Details}
We use the cross-entropy loss for both classification tasks. Both networks use leaky-ReLU nonlinearities \cite{maas2013rectifier}, max-pooling for downsampling, and batch normalization layers for stabilization and regularization \cite{ioffe2015batch}. To improve the generalization of our model, we perform several data augmentations during training. Details of the architecture and training can be found in the supplementary document.

\section{Results and Analysis} \label{section:results}

We analyze the classification accuracy of each model evaluated on scenes that were not present in the training set. We formulate a simplified theoretical model to study the behavior of the signal-to-noise ratio (SNR) as a function of scene parameters, and estimate the empirical SNR in our data. Furthermore, we use a controllable synthetic 2D data generation setup to study the one vs. two classification accuracy for different relative motions of two people.

\subsection{Data Collection}
We collected a dataset of 30 scenes per task in interior spaces such as offices, conference rooms, and public lounges. This dataset was split into training, validation and test sets of 20, 5, and 5 scenes to study the generalization of the models on new unseen scenes. The dataset consists of total of 12 hours of video for each task, split equally across all classes. All videos of the blank walls were recorded in 16-bit format at 15 frames per second using a PointGrey Grasshopper 3 camera. The subjects were 2-5 meters away from the observed wall. We ensured that the subjects did not cast any direct shadow on the wall that would be visible to the naked eye. No additional lights were placed to illuminate the scene or to increase the signal reflected off the subjects. The exposure parameters were adjusted based on the ambient light in the given environment.

\textbf{Classification of number of people.} The subjects were instructed to occasionally vary their walking speeds and trajectories. In the two person scenarios, the subjects aimed for independent movements, but occasionally moved in lockstep or otherwise mutually coordinated patterns.

\textbf{Activity recognition task.} Each activity was performed by one person in the hidden scene, as done in other activity recognition datasets~\cite{soomro2012ucf101}. Similar to the number of people classification task, the subjects were asked to vary their speed and trajectory for walking. For the jumping task, the subjects varied frequencies in different locations of the hidden scene and changed their location randomly. When collecting the data for crouching, the subjects started with walking then crouched after some time. We believe that this activity is the closest simulation of falling while walking. For the waving hands activity class, the subjects randomly moved their hands while standing, and switched their location in the hidden scene at random intervals.

\subsection{Classification Results}
The networks were evaluated on unseen data from the separate set of test scenes, with randomly extracted 256-frame (17 second) segments. We achieve an accuracy of 94.4\% in classifying the number of people and 93.7\% in the activity recognition task. To verify that the models are also useful with shorter video segments, we additionally trained and tested models using 64 and 128 frame inputs, which led to a modest loss of accuracy (refer to Table \ref{table:accuracy}). The remainder of the analysis was done with the 256-frame model.

\begin{table}[t]
\centering
\resizebox{\columnwidth}{!}{
\begin{tabular}{|c|c|c|c|c|}
\hline
\textbf{Task} & \textbf{Frames} & \textbf{Train} & \textbf{Val} & \textbf{Test} \\ \hline
\multirow{3}{*}{Classifying number of people} & 64 & \begin{tabular}[c]{@{}c@{}}86.8\%\end{tabular} & 82.4\% & 79.4\% \\ 
 & 128 & 92.9\% & 90.8\% & 88.3\% \\
 & 256 & 96.4\% & 95.8\% & 94.4\% \\ \hline
\multirow{3}{*}{\begin{tabular}[c]{@{}c@{}}Activity recognition\end{tabular}}  & 64 & 91.5\% & 90.8\% & 88\% \\ 
 & 128 & 94.6\% & 91.4\% & 90.8\% \\ 
 & 256 & 96.3\% & 94.4\% & 93.7\% \\ \hline
\end{tabular}
}
\caption{Accuracy for classifying number of people and activity recognition tasks.}
\vspace{-2em}
\label{table:accuracy}
\end{table}

\subsection{Real Time Application}
We packaged our trained models into a real time system for in-situ deployment. To estimate the mean frame to be subtracted, we maintain an average of all frames observed during the session. The estimated mean stabilizes into a useful estimate within the first 10 seconds. Our system outputs new predictions 15 times per second, based on a 17-second window of most recent frames. We run it on a 6-core laptop using a Nvidia RTX 2080 Max-Q GPU. Our model generalizes to new scenes when tested using the real time system. We also found that the system worked well in 10 outdoor scenarios, despite the absence of any outdoor configuration in the training data. The outdoor setups were similar to Figure \ref{fig:setup} except that there was no wall hiding the unknown area but the camera only observed the blank wall. Illumination came from the sun or street lamps. Our model was robust to small temporal lighting fluctuations caused by changes in position of the sun and movement of clouds. Refer to the supplementary video for the real time demo.

\subsection{Limitations}
\begin{figure}[t]
\centering
\subfloat[\label{mat:confmat_number}]{%
\includegraphics[width=0.43\columnwidth]{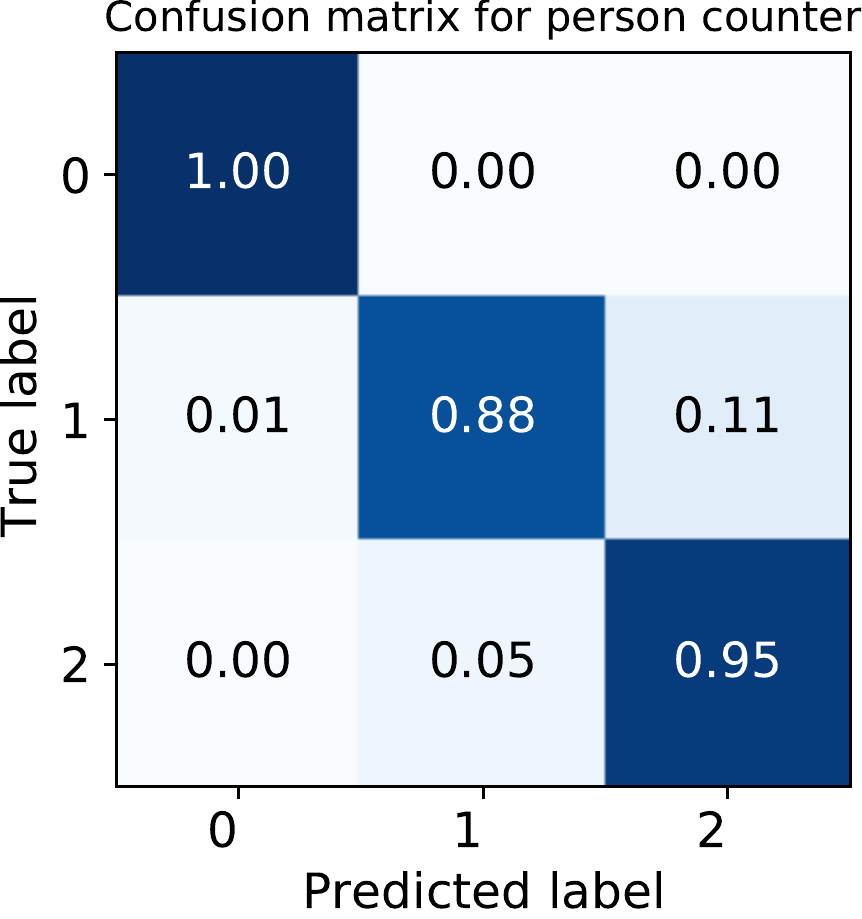}}
\hfill
\subfloat[\label{mat:confmat_actrecog}]{%
\includegraphics[width=0.53\columnwidth]{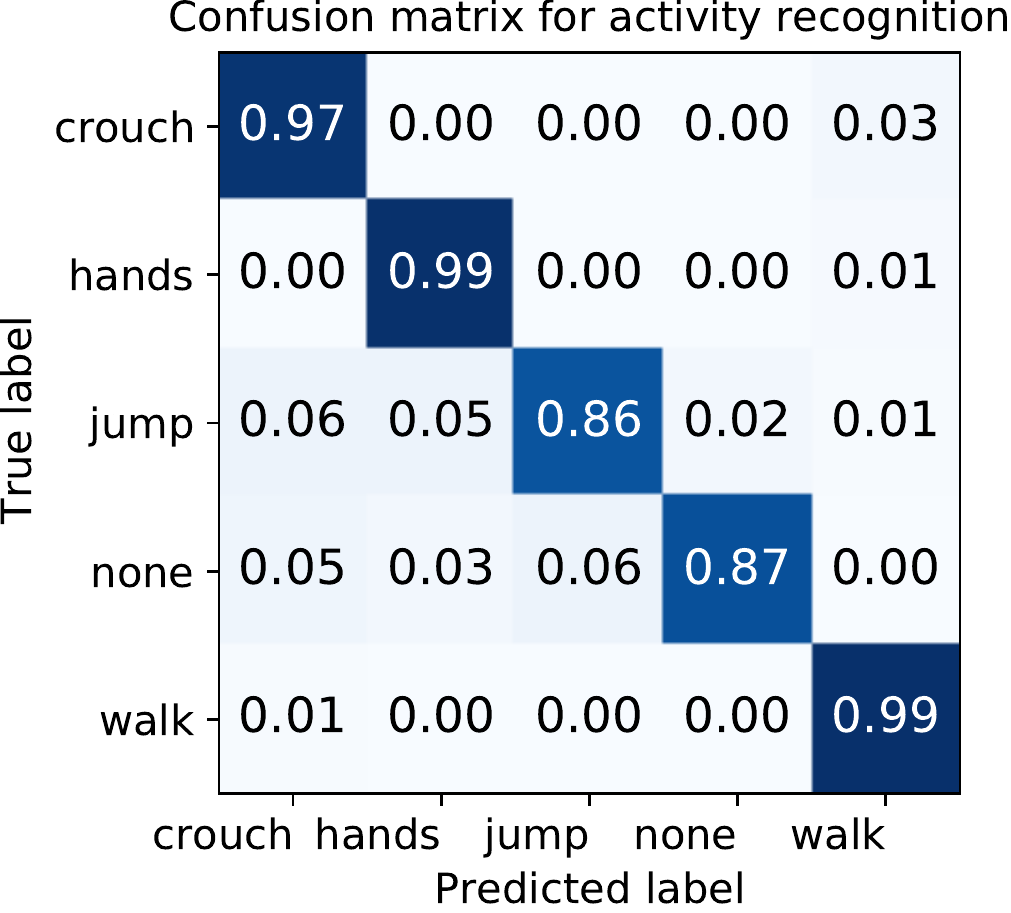}
`}
\caption{Confusion matrix on the test set comprising of 5 unseen test scenes for both tasks, (a) classifying number of people and (b) activity recognition task.}
\vspace{-1em}
\label{fig:confusion}
\end{figure}
\begin{figure}[t]
\centering
\includegraphics[width=0.7\columnwidth]{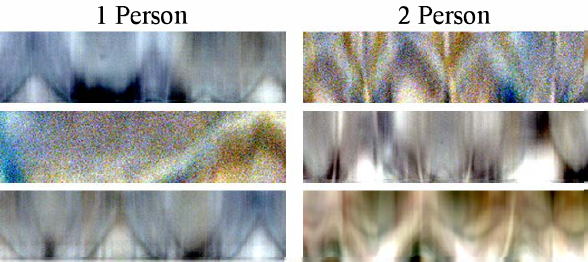}
\caption{Misclassified sample plots for one and two person class predicted as the other class.}
\vspace{-1em}
\label{fig:misclassify}
\end{figure}

The performance of our models suffers in very low light conditions, and under irregular lighting variations such as a television illuminating the scene. Other difficult scenarios are conditions with low contrast in the persons’ clothing and the background, high distance of the subjects to the observed wall, and lockstep motion of two persons.

Figure \ref{fig:confusion} shows confusion matrices for the two tasks to reveal the performance of the models on individual classes.
For classifying the number of people, we observe that the model is most accurate for the zero person case, as it is the easiest class. Our model sometimes misclassifies between the one and two person class, with better accuracy on the two-person class. On visual inspection of the space-time plots and the ground truth, we observe that the misclassified one-person samples share a low contrast signal (e.g. the person was very far from the observed wall, or was wearing clothes of matching color as the background wall). In the misclassified two person samples, we observe a trend that the two people walked in lockstep, resulting in visually similar space-time plots to a one person case. These trends can be seen in the samples shown in Figure \ref{fig:misclassify}. These two observations motivated analysis about the impact of these properties, which we present in subsections~\ref{section:snr} and~\ref{section:analysis_motion}.

For the activity recognition task (see Figure \ref{mat:confmat_actrecog}), the highest misclassification error between jumping and no activity can be attributed to samples where the subject jumped with high frequency and moved less along both spatial dimensions. Such a motion does not result in meaningful signal in either the horizontal or vertical space-time plot.

\subsection{Analysis of the Signal-to-Noise Ratio}\label{section:snr}
To understand the impact of scene parameters on the signal quality, we derive a simplified mathematical model for our measurements under idealized conditions. Main results are presented here; refer to supplementary document for the derivations. We then relate this analysis to observed SNR in our real-world data. 

\paragraph{Estimated Signal Power.} Consider an idealized scene, where we are observing a diffuse wall of albedo (color) $\alpha$, surrounded by a static and constant-colored environment where the incident radiance is $L_s$ (i.e., the intensity that a camera would record when photographing the surrounding scene). A person is facing the wall at distance $d$, occupying an area $A = w \times h$ (as projected on the wall). Let the radiance of the person be $L_p$. 

The radiance recorded on the wall can then be computed using the \emph{rendering equation} \cite{Kajiya:1986:RE:15922.15902}. In the supplementary document, we derive the expected power of the observed radiance signal after subtracting the mean to be
\begin{equation}
P \approx \left[ \frac{\alpha A \sqrt{q}}{\pi d^2} (L_p - L_s) \right]^2,
\end{equation}
where $q \in [0,1]$ is the fraction of the time that the person is near the wall.

\paragraph{Estimated Noise Power.}
The camera photon noise at a pixel observing radiance $\alpha L_s$ (the average observed color of the wall) is approximately normally distributed with variance $k \alpha L_s$. The value of $k$ depends on the camera exposure settings, and is in fact quadratically proportional to the gain \cite{hasinoff2010noise}. Furthermore, when $s$ pixels are averaged into one (e.g. when an image is downsampled or projected into a space-time plot), the variance reduces to $k \alpha L_s / s$.

\paragraph{Signal to Noise Ratio.}
The ratio of the power of the signal to the variance of the noise in decibels is
\begin{eqnarray}\label{eq:SNR}
    10 ~ \mathrm{log}_{10} \left[ \frac{A \sqrt{q \alpha s L_s} (C - 1)}{\pi d^2 \sqrt{k}}  \right]^2,
\end{eqnarray}
where $C = L_p / L_s$ is the contrast ratio between the color reflected off the person and the environment.

In this formula, we can identify various relationships between the SNR and the scene parameters. As expected, subjects occupying larger area generate a stronger signal. The SNR has a strong inverse dependence to the distance of the person, making it difficult to distinguish distant persons in videos and space-time plots. Low contrast between the person and the background (e.g. white clothes against a white background) also contributes to a low SNR, as do low light conditions and various wall surface materials. Attempts to brighten the observed signal by increasing camera gain are counteracted by a proportional boost to the noise level. The SNR can be improved by averaging the image over multiple pixels by downsampling along the axes at a trade-off with the spatial sharpness of the signal. 

Randomly sampling the scene parameters within the range of plausible values, our model predicts an average SNR of $-20 ~\mathrm{dB}$, with standard deviation of $10~ \mathrm{dB}$.

\begin{figure}[t]
\centering
\includegraphics[width=\columnwidth]{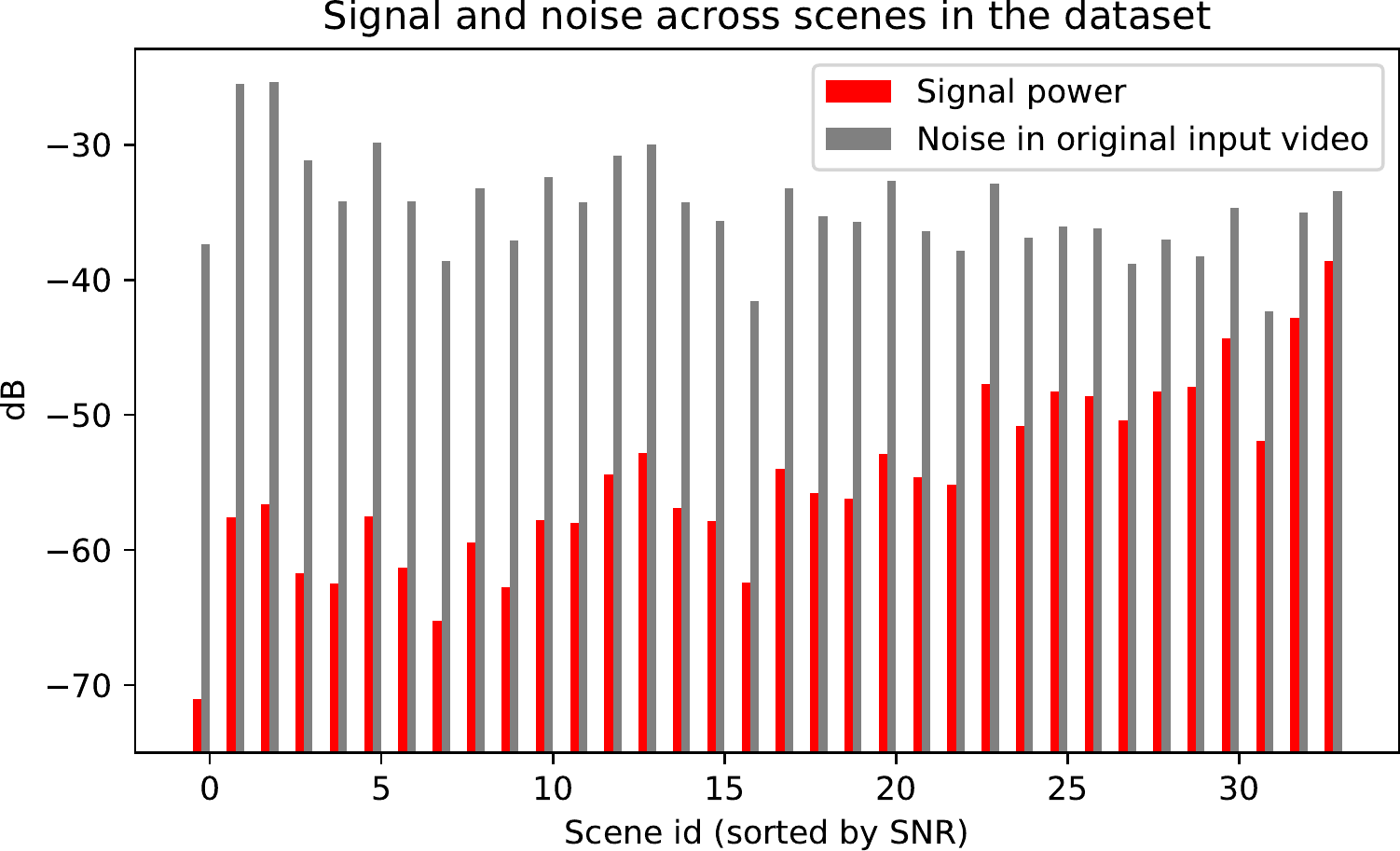}
\caption{The estimated signal and noise power in our datasets. The values correspond to full resolution frames. The signal-to-noise ratio is the difference between the signal and noise decibels, ranging from roughly $-5$ to $-35~\mathrm{dB}$.}
\vspace{-1em}
\label{fig:snrplot}
\end{figure}

\paragraph{SNR in Real Data.}
A video $V$ (with zero mean over time) can be viewed as a sum of signal ($S$) and noise ($N$) videos, $V = S+N$. In the supplementary document, we show that the power (i.e. variance) of each component in our data can be estimated as $\mathrm{Var} ~ N \approx \frac{1}{2} \mathrm{Var} ~ (V - \mathcal{T}(V))$ and $\mathrm{Var}~S \approx \mathrm{Var} ~ V - \frac{1}{2} \mathrm{Var} ~ (V - \mathcal{T}(V))$, where $\mathcal{T}$ is an operator that shifts the video temporally by one frame.

In Figure \ref{fig:snrplot}, we plot the signal and noise power from these estimates for all of our training and test datasets. We show the noise level corresponding to the original full-frame resolution. The noise is significantly stronger than the signal in all datasets. The signal-to-noise ratios in the real data range from $-5$ to $-35 ~\mathrm{dB}$, which roughly agrees with the predictions of the mathematical model. Downsampling and collapsing the videos to space-time plots corresponds to averaging $s=30,000$ pixels into every pixel in the plot, yielding a noise reduction of approximately $45~ \mathrm{dB}$.

\begin{figure}[t]
\centering
\subfloat{%
\includegraphics[width=0.5\columnwidth]{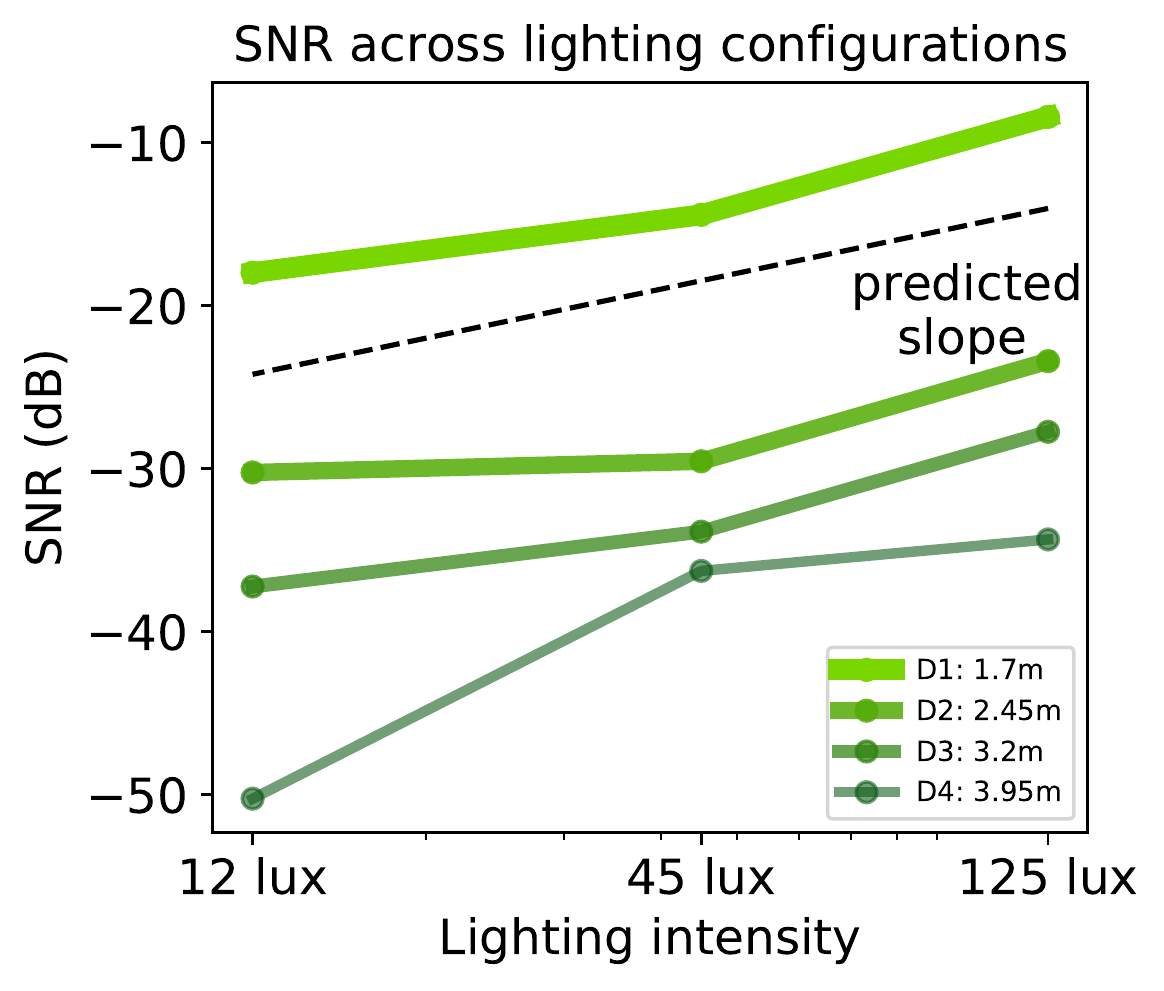}
}
\subfloat{%
\includegraphics[width=0.5\columnwidth]{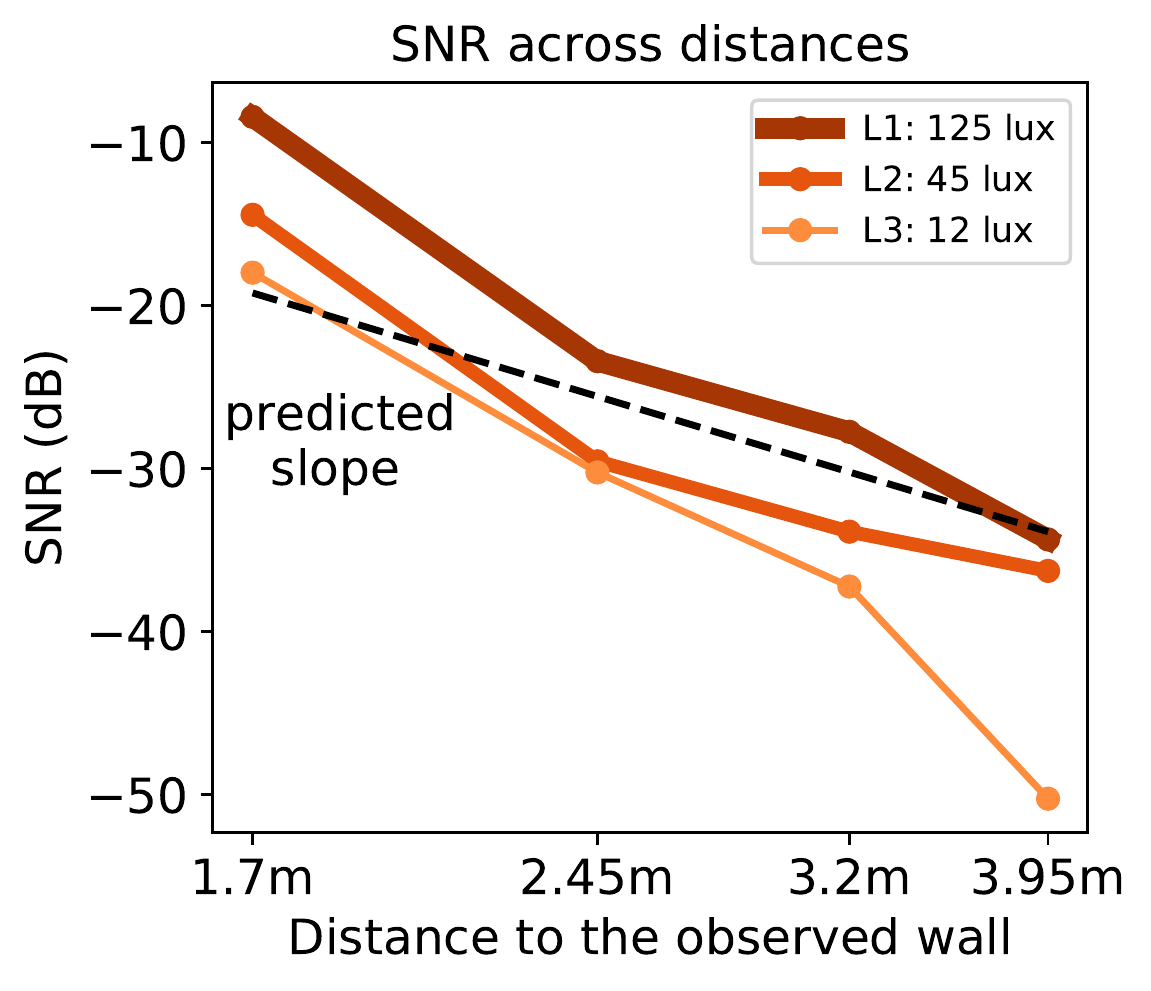}
}
\caption{Empirical SNR as a function of lighting intensity and subject distance. The slopes of the dotted lines show the approximate theoretical prediction of the dependency based on Equation \ref{eq:SNR}.}
\vspace{-1em}
\label{fig:snr_distance_light}
\end{figure}

In addition to our dataset, we captured data controlling the subject distance to the observed wall and the light intensity in the room. We conducted all activities with a single person in an unseen room at 4 distances (1.7, 2.45, 3.2, and 3.95 metres) and 3 lighting intensities (12, 45, and 125 lux). We gathered data for 4 activities (walking, waving hands, crouching and jumping), with at least 3 segments for each combination of activity, lighting, and distance. 

Figure \ref{fig:snr_distance_light} plots the SNR as a function of distance and light configurations. 
The slopes of the curves are consistent with the approximate theoretical prediction. The slight divergence of the plots is due to the $1/d^2$ term being an approximation in Equation \ref{eq:SNR} that ignores the global illumination effects due to multiple bounces. Such effects can be particularly important in small enclosed spaces like our room. Refer to the supplementary document for further details.

\subsection{Analysis of Human Motion with Synthetic Data}\label{section:analysis_motion}
For classification of the number of people, the performance also strongly depends on the relative motion of the people. To experiment with classification accuracy under controlled motion of the people in the hidden scene, we formulated a simple synthetic data generation pipeline that simulates our imaging setup in a simplified 2D ``flatland'' scenario. Details of the pipeline are in the supplementary document. To gauge the realism of the synthetic data, we observe that our classifier trained with the real data achieves an overall accuracy of $71.5\%$ in one-vs-two classification for the synthetic test dataset of 14,000 samples.

\begin{figure}[t]
    \centering
     \subfloat[\label{fig:synthanalysis1}]{
      \includegraphics[width=0.45\columnwidth]{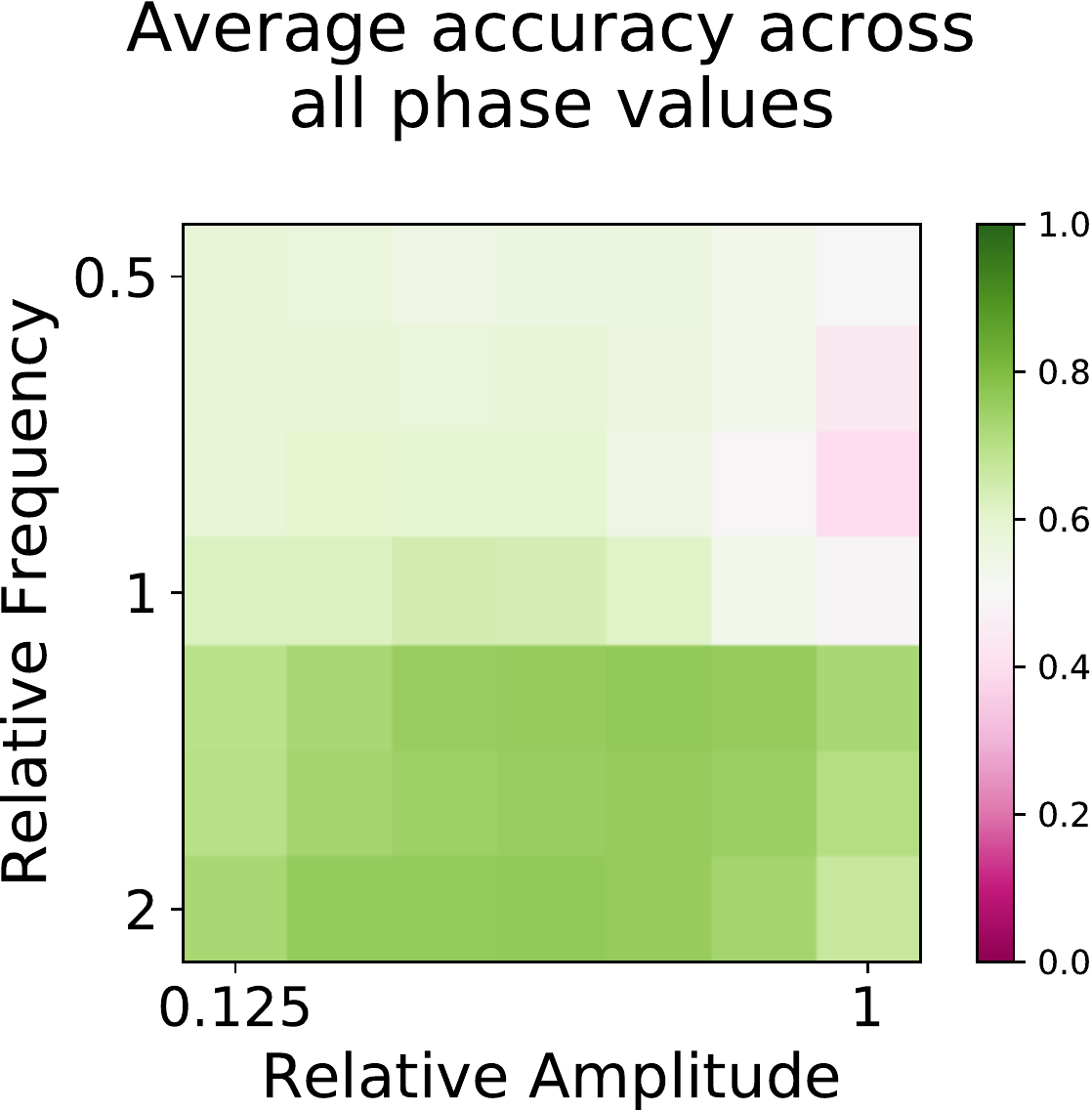}}
     \subfloat[\label{fig:synthanalysis2}]{
      \includegraphics[width=0.45\columnwidth]{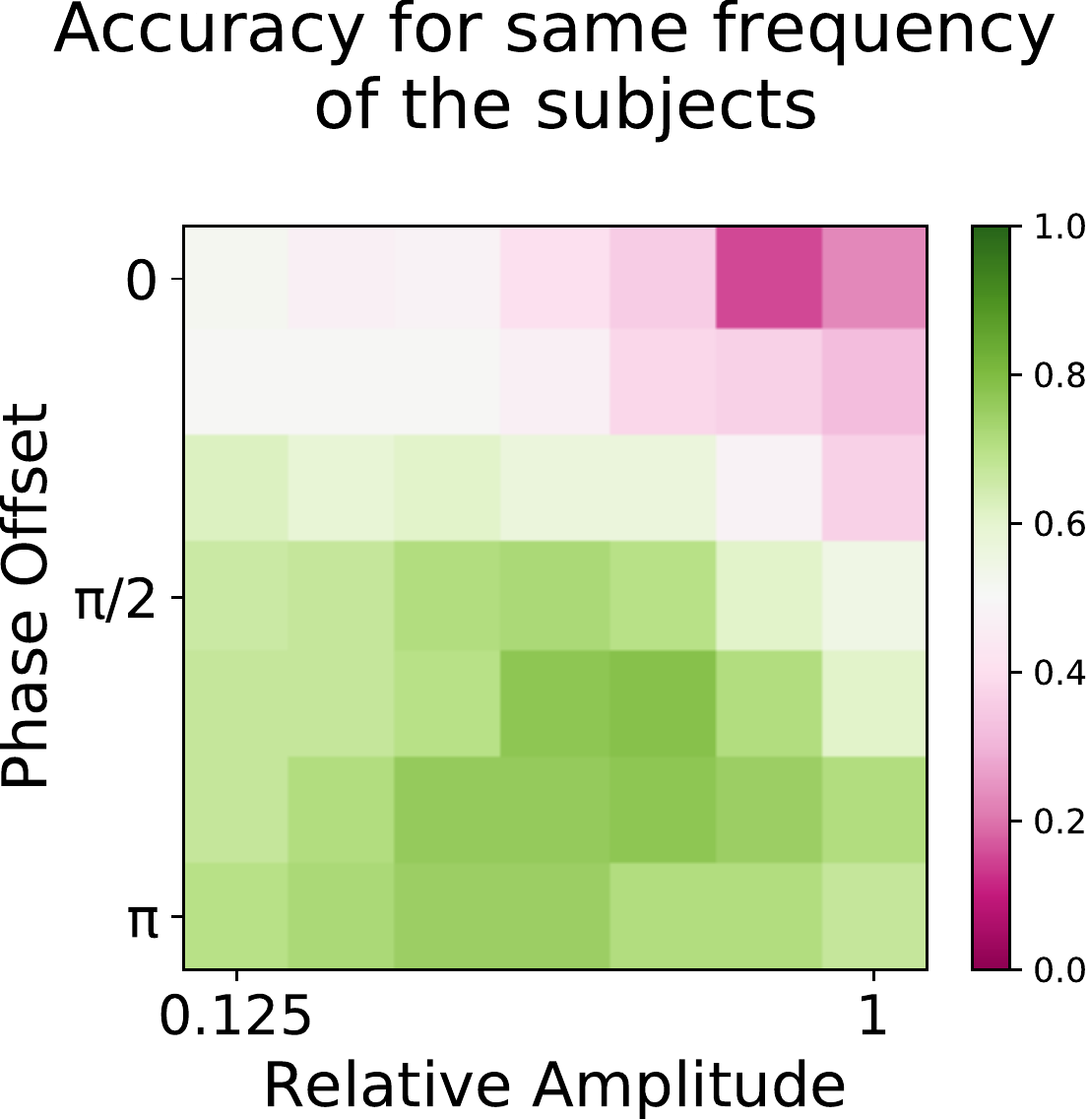}
     }
    \qquad
     \subfloat[\label{fig:synthplot2}]{%
      \includegraphics[width=0.5\columnwidth]{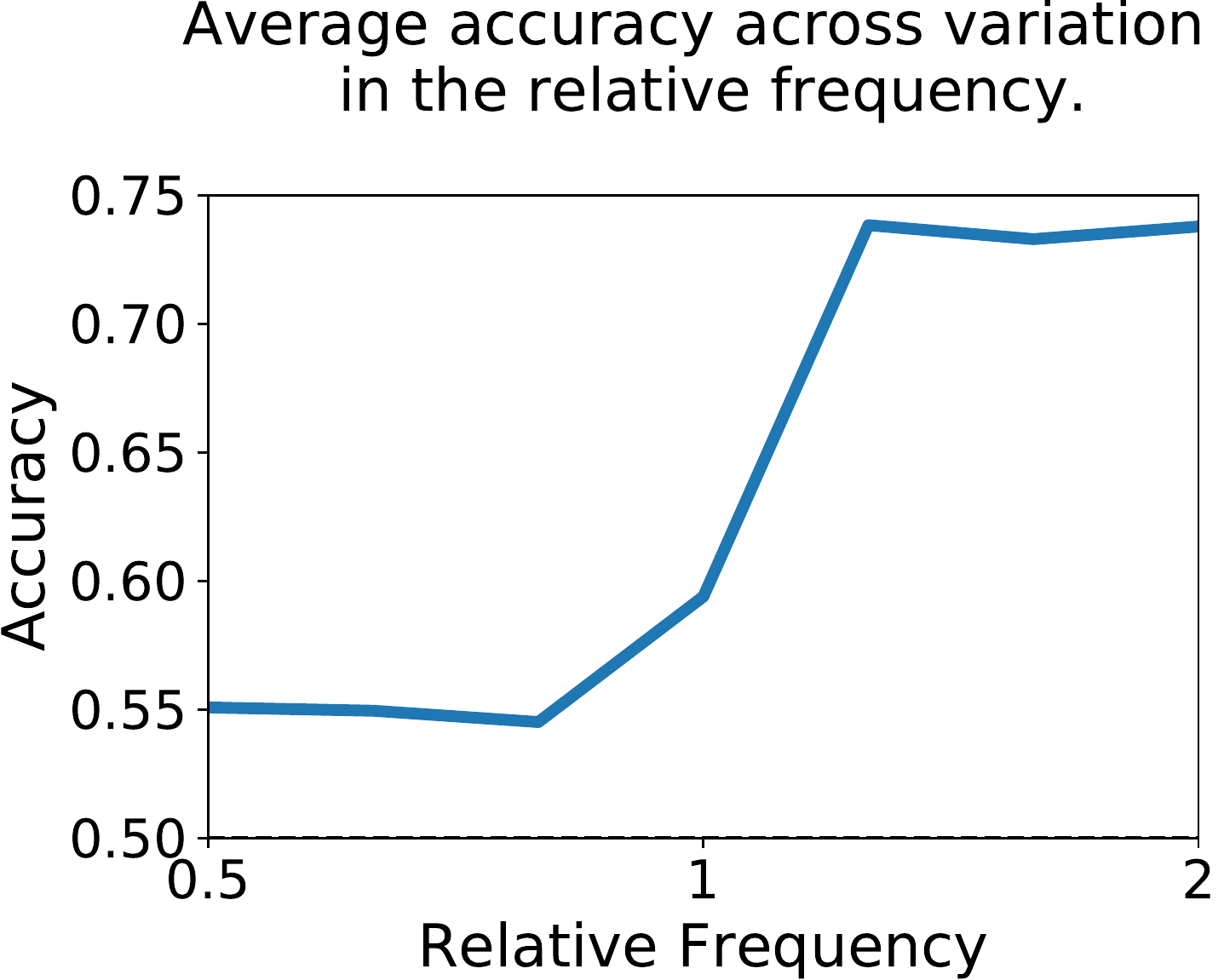}
     }  
     \subfloat[\label{fig:synthplot1}]{%
      \includegraphics[width=0.5\columnwidth]{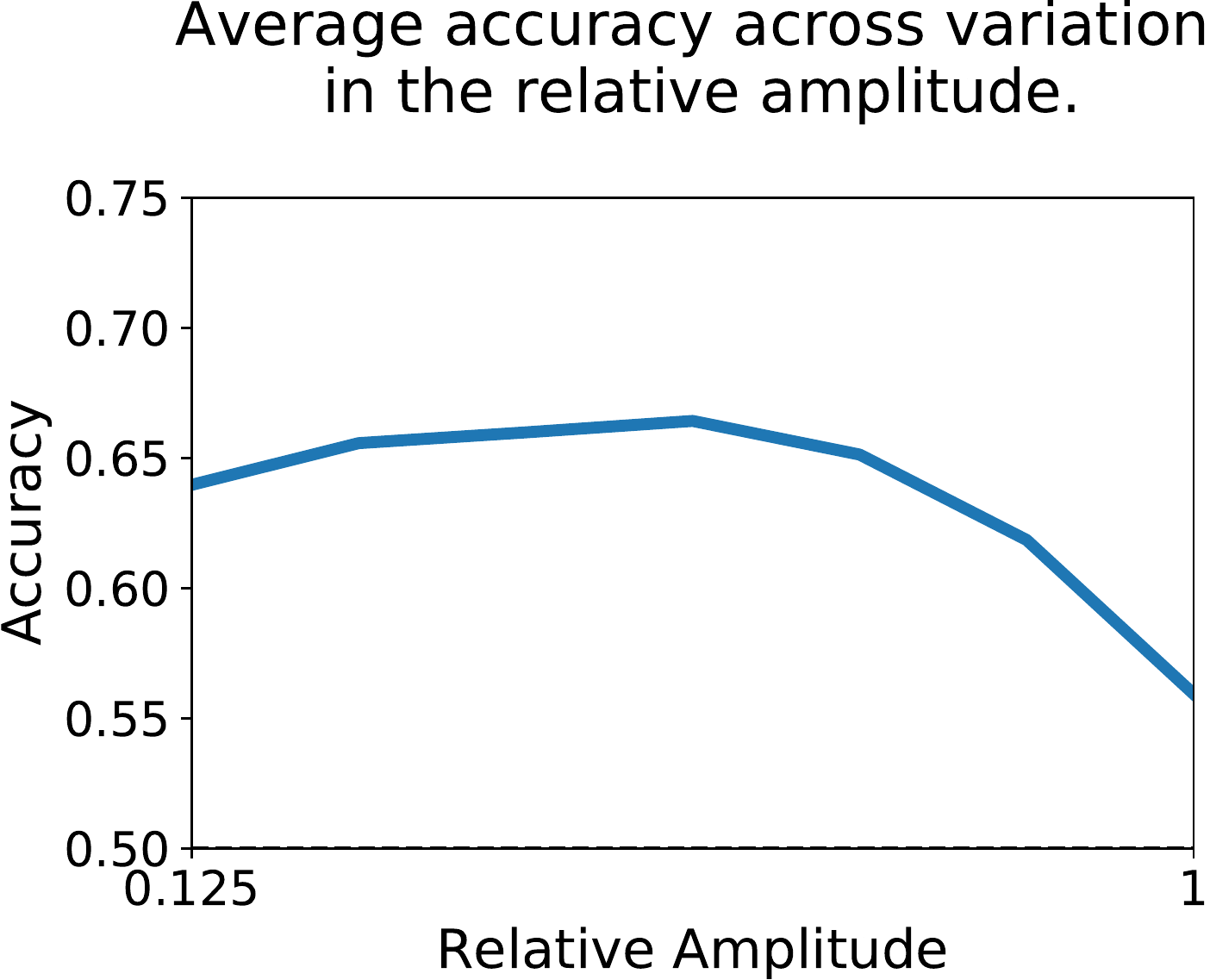}
     }
\caption{Classification performance for synthetic two-person video segments as a function of different relative motion parameters.}
\vspace{-1em}
\label{fig:synthanalysis}
\end{figure}

To study the impact of relative motion, we generated a dataset of two persons moving in various parameterized sinusoidal patterns under 80 different scenes, and studied the average classification performance (i.e. if the segments were correctly classified as two person cases) as the relative motion was varied. Specifically, we set person 1 to move in a standard sinusoidal pattern, and varied the relative frequency, amplitude, and phase of the second person's movements. Figure \ref{fig:synthanalysis} shows the results of this analysis along different axes of variation.

Figure \ref{fig:synthanalysis1} shows the effect of varying the relative amplitude and frequency of the second person's movement. The lowest classification accuracy is attained when the second person's motion is both slow and of equal amplitude with person 1. The overall accuracy improves as the relative motion frequency increases, as shown in Figure \ref{fig:synthplot2}, confirming the hypothesis that plots with high frequency motion lead to better classification. The accuracy also generally improves when the amplitudes of the motions are different (Figure \ref{fig:synthplot1}); it appears that having contrast between the motions is more valuable than simply having large motions. Indeed, even motions with small relative amplitude appear to be sufficient for good classification.

Figure \ref{fig:synthanalysis2} analyzes the special case when both persons are moving at the same frequency, but their relative phase and amplitude are varied. As expected, the classification accuracy is low when the persons move in lockstep, i.e. when the phase difference is small. The lowest accuracy is achieved when the motions are of equal amplitude, namely identical. In contrast, opposite-phase and equal magnitude motion appears to be resolved best -- because the signals from the people are maximally different while being least likely to hide one another.
The generally successful classification for equal-frequency cases shows that accuracy is not affected when the two person data is a one-dimensional manifold, ruling out that the network would be simply performing a dimensionality analysis of the data. This suggests that the model in fact uses visual features such as texture differences in the input space-time plots.

\section{Conclusion}

We show that a non-line-of-sight setup with no prior calibration or knowledge of an occluder can recover meaningful information about activity in a hidden scene. The motion of the persons blocks and reflects faint indirect light, producing signals on the observation plane that reveal the dynamic activity in the hidden scene.
We amplify and process this signal into a space-time plot that we analyze using a convolutional neural network. We demonstrate automatic classification between zero, one, or two moving person(s), and infer the activity of one person.
These models are trained on a dataset of 20 different scenes. Once trained, the method achieves high accuracy both in unseen test data and real-time in-situ deployment.

We studied aspects of the imaging setup and the task, analyzing the signal and noise in the measurements with a mathematical model, and the effect of motion patterns on the identifiability of the number of people using synthetic data. We find that factors such as distance of the person to the observed wall, relative contrast of the person to the background wall, and relative motion in the case of two people affect the observed signal.

\textbf{Acknowledgements.} This work is supported by NSF under Cooperative Agreement PHY-2019786 (IAIFI), NSF Award 1955864, NSF Grant No. CCF-1816209, and DARPA REVEAL Program under Contract No. HR0011-16-C-0030. We would like to thank Jeffrey Shapiro, Franco Wang, and Vivek Goyal for helpful discussions, Luke Anderson and Parimarjan Negi for proofreading, and Harshvardhan and Samip Jain for their help in data collection.

{\small
\bibliographystyle{ieee_fullname}
\bibliography{main}
}

\end{document}